\documentclass[runningheads]{llncs}

\usepackage{eccv}

\usepackage{eccvabbrv}

\usepackage{graphicx}
\usepackage{booktabs}

\usepackage{subcaption}

\newcommand{\bfx}[0]{\mathbf{x}}
\newcommand{\bfy}[0]{\mathbf{y}}
\newcommand{\bfz}[0]{\mathbf{z}}

\newcommand{\bfv}[0]{\mathbf{v}}

\usepackage{caption} 
\usepackage[accsupp]{axessibility}  %

\usepackage{hyperref}

\usepackage{orcidlink}

\begin{document}

\title{Adversarial Attacks on Hyperbolic Networks}

\author{Max van Spengler\inst{1}\orcidlink{0000-0002-7440-920X} \and
Jan Zah\'{a}lka\inst{2}\orcidlink{0000-0002-6743-3607} \and
Pascal Mettes\inst{1}\orcidlink{0000--0001-9275-5942}}

\authorrunning{M. van Spengler et al.}

\institute{University of Amsterdam, the Netherlands \and
Czech Technical University in Prague, Czech Republic\\
\email{jan.zahalka@cvut.cz}}

\maketitle

\begin{abstract}
As hyperbolic deep learning grows in popularity, so does the need for adversarial robustness in the context of such a non-Euclidean geometry. To this end, this paper proposes hyperbolic alternatives to the commonly used FGM and PGD adversarial attacks. Through interpretable synthetic benchmarks and experiments on existing datasets, we show how the existing and newly proposed attacks differ. Moreover, we investigate the differences in adversarial robustness between Euclidean and fully hyperbolic networks. We find that these networks suffer from different types of vulnerabilities and that the newly proposed hyperbolic attacks cannot address these differences. Therefore, we conclude that the shifts in adversarial robustness are due to the models learning distinct patterns resulting from their different geometries.
\end{abstract}

\section{Introduction}
\label{sec:introduction}

Non-Euclidean deep learning is a rapidly growing field where various inductive biases are instilled into our models through the use of geometric techniques \cite{bronstein2017geometric}. For example, many works focus on challenging the usual choice of Euclidean space as the foundation for our models by replacing it with various non-Euclidean geometries, such as hyperspherical \cite{liu2017deep,kasarla2022maximum} or hyperbolic \cite{ganea2018,shimizu2020} spaces. Such geometries allow us to shed new light on deep representation learning. Each comes with its own benefits and drawbacks, highlighting the importance of judiciously choosing which space to learn in.

Hyperbolic geometry specifically has shown to significantly improve the representational capacity of models when dealing with hierarchically structured data \cite{nickel2017poincare}. In many areas of machine learning, from natural language \cite{tifrea2018poincar,chen2021fully,zhu2020hypertext} and graph learning \cite{chami2019hyperbolic,dai2021hyperbolic,liu2019hyperbolic,xue2024residual,zhang2021hyperbolic} to computer vision \cite{khrulkov2020hyperbolic,ghadimi2021hyperbolic,ermolov2022hyperbolic,spengler2023}, hierarchical structures are ubiquitous. Consequently, many existing architectures benefit from incorporating hyperbolic geometry into some or all of their layers. As the importance of hyperbolic deep learning grows, so does the need for adversarial robustness of models adopting it. Early evidence \cite{spengler2023} has shown that adversarial attacks designed for Euclidean models have degraded performance against hyperbolic models, underlining the need for adversarial attacks that operate in the same geometry in which the networks are trained.

As applications of Artificial Intelligence are swiftly permeating everyday life, we are becoming increasingly susceptible to malign adversaries through, often unknown, weaknesses in our models. This has led to a surge in interest for universally applicable adversarial attacks that can be used to pinpoint and quantify any such underlying vulnerabilities \cite{goodfellow15fgsm,szegedy14advpert}. In the absence of known powerful attacks, models can neither be evaluated for adversarial robustness nor protected against malicious intent \cite{goodfellow15fgsm,wong2020fast}.

Over the years, several well performing universal adversarial attack frameworks have been proposed for various modalities, such as AutoAttack \cite{croce2020autoattack} for images and TextAttack for text \cite{morris2020textattack}. However, existing adversarial attacks are designed against networks built in Euclidean space and their performance for other geometries cannot be guaranteed. In fact, models with different geometries can be more robust to these universal attacks \cite{spengler2023}, suggesting that the existing methods are inappropriate for non-Euclidean settings.
We therefore require universal adversarial attacks that are geometry-agnostic in nature.
The aim of this paper is to make a first step towards generalized attacks and to provide some insights into their behaviour when applied to Euclidean and hyperbolic networks.

In this paper, we propose Riemannian generalizations of the FGM \cite{goodfellow15fgsm} and PGD \cite{madry18pgd} adversarial attacks, with a specific application for hyperbolic networks.
First, we use the proposed attacks on synthetic datasets to showcase the differences between the geometry-informed attacks and existing Euclidean attacks. Second, we investigate standard and our Riemannian adversarial attacks on identically trained Euclidean and hyperbolic ResNets to understand their effect and the differences in behaviour. With the introduced geometry-agnostic adversarial attacks and the empirical results, we hope to make a first step towards adversarially robust geometric deep learning and to shine some light on the potential difficulties that will have to be solved along the way.

\section{Related work}
\label{sec:related_work}

\subsection{Adversarial attacks}

Adversarial attacks change a model's output by adding subtle adversarial perturbations to the input data \cite{goodfellow15fgsm,szegedy14advpert}. The most common type of adversarial perturbation is noise restricted in terms of a norm: usually $\ell^\infty$ \cite{goodfellow15fgsm,kurakin17pgd,madry18pgd}, but $\ell^0$ \cite{williams2023sparse} and $\ell^2$ \cite{carlini2017cwattack} norm attacks are viable as well. Beyond adversarial noise, adversarial patches \cite{hu21patch,walmer22patch}, frequency perturbations \cite{feng22freq}, camouflage \cite{feng23camo,hu23camo}, shadow \cite{li23shadow, zhong22shadow}, rain \cite{yu22rain}, or even 3D meshes \cite{yang2023mesh} may be used as adversarial perturbations. Adversarial attacks can be categorized into white and black-box. White box attacks have full access to the model and leverage the model's gradient to optimize the adversarial perturbation. The most widely used general attacks are the fast gradient sign method (FGSM) \cite{goodfellow15fgsm} and projected gradient descent (PGD) \cite{kurakin17pgd,madry18pgd}. Black box attacks optimize perturbations solely based on the model output, either on output probabilities \cite{andriushchenko2020square} or hard labels \cite{kahla2022label}. Overall, there is a wide variety of adversarial attacks, and they boast a high attack success rate. This presents a security problem, since a malicious attacker has a wide array of tools to exploit machine learning models.

The weakness to adversarial attacks is innate to modern models. To safeguard a model against adversarial attacks, one needs to employ adversarial defense techniques. The most widely used approach is adversarial training (AT) \cite{goodfellow15fgsm,madry18pgd}: add adversarially perturbed images with the correct labels to the training dataset, so that the model knows how to classify them correctly. AT does improve robustness, but may reduce accuracy on clean data and incurs additional computational costs. Alternative defense techniques include adversarial distillation \cite{papernot2016distillation}, attack detection \cite{pang2018advex,xu2018advex}, and discarding \cite{bai2021improving,xie2019feature,yan2021cifs} or recomputing \cite{kim2023feature} the features compromised by the attacks. Like in classic cybersecurity, a cat-and-mouse game is on in machine learning security: adversarial defense keeps catching up with the adversarial attacks, but never catches up completely. With the increased adoption of machine learning models in applied practice, attacks are increasingly lucrative to the attackers. Therefore, model stakeholders cannot afford to leave them undefended. Hyperbolic models are no exception, and the time is ripe to investigate adversarial defenses for them as well.

\subsection{Hyperbolic learning}
The field of hyperbolic learning forms a recent and fast-growing research direction in deep learning \cite{mettes2024hyperbolic,peng2021hyperbolic,zhou2023hyperbolic}, focusing on addressing known limitations of using Euclidean geometry as the geometric basis, including but not limited to hierarchical modelling, low-dimensional learning, and robustness. Early work showed that embedding hierarchies in hyperbolic space is vastly superior to Euclidean space \cite{ganea2018hyperbolic,law2019lorentzian,nickel2017poincare}, allowing for hierarchical embeddings with minimal distortion \cite{sarkar2011low}.
The results on hyperbolic embeddings of hierarchies sparked great interest in deep learning with hyperbolic embeddings or hyperbolic layers. As a natural extension to hierarchies, a wide range of works investigated hyperbolic graph learning \cite{chami2019hyperbolic,dai2021hyperbolic,liu2019hyperbolic,xue2024residual,zhang2021hyperbolic}, with applications in molecule classification \cite{zhang2021lorentzian}, recommendation \cite{sun2021hgcf,wang2021fully,yang2023hyperbolic}, and knowledge graph reasoning \cite{wang2023hygge}. Hyperbolic learning has furthermore been shown to be an effective geometric network basis for text embeddings \cite{tifrea2018poincar}, speaker identification \cite{lee2021lightweight}, single-cell embeddings \cite{klimovskaia2020poincare}, traditional classifiers \cite{cho2019large,doorenbos2043hyperbolic}, reinforcement learning \cite{cetin2023hyperbolic} and more.

Within computer vision, early works showed the latent hyperbolic nature of visual datasets \cite{khrulkov2020hyperbolic} and the potential of hierarchical classification with hyperbolic embeddings \cite{dhall2020hierarchical,liu2020hyperbolic,long2020searching}. Visual learning with hyperbolic embeddings has since shown to improve a wide range of problems in computer vision, such as semantic segmentation \cite{atigh2022hyperbolic,franco2024hyperbolic,kwon2024improving,weber2024flattening,weng2021unsupervised}, fine-grained classification \cite{ermolov2022hyperbolic,ghadimi2021hyperbolic}, few-shot learning \cite{gao2021curvature,guo2022clipped}, video and temporal learning \cite{franco2023hyperbolic,li2024isolated,suris2021learning}, vision-language learning \cite{desai2023hyperbolic,kong2024hyperbolic,ramasinghe2024accept}, and generative learning \cite{bose2020latent,hsu2021capturing,mathieu2019continuous,nagano2019}. Based on these advances, recent works have made the step towards fully hyperbolic visual representation learning, where all network layers are embedded in hyperbolic space \cite{spengler2023,bdeir2023fully}. Fully hyperbolic convolutional networks have shown to improve out-of-distribution robustness and increased adversarial robustness. Such evaluations were however performed using conventional adversarial attacks, which assume a Euclidean geometry as basis. The introduction of fully hyperbolic networks requires adversarial attack algorithms in hyperbolic space. In this paper, we show how to generalize well-known attacks such as FGSM and PGD to hyperbolic space.
\section{Background}
\label{sec:background}
This paper will use the Poincaré ball model of hyperbolic space. Here, we provide a short overview of the formulas that will be of importance in later sections. For a more thorough treatment of hyperbolic space and the different isometric models that are used to study it, the reader is referred to \cite{cannon1997,anderson2006}. The Poincaré ball model is defined as the Riemannian manifold $(\mathbb{D}_c^n, \mathfrak{g}_c)$, where $-c$ is the constant negative curvature of the corresponding hyperbolic space. The manifold is defined as
\begin{equation}
    \mathbb{D}_c^n = \{\bfx \in \mathbb{R}^n : ||\bfx||^2 < \frac{1}{c}\},
\end{equation}
and the Riemannian metric as
\begin{equation}
    \mathfrak{g}_c = \lambda_\bfx^c I_n, \quad \lambda_\bfx^c = \frac{2}{1 - c ||\bfx||^2},
\end{equation}
where $I_n$ denotes the $n$-dimensional identity matrix.

There exists no binary operation on this manifold that could be used to turn it into a vector space. However, we can use the non-commutative Möbius addition operation to get a similar but slightly weaker gyrovector structure \cite{ungar2022}
\begin{equation}
    \bfx \oplus_c \bfy = \frac{(1 + 2 c \langle \bfx, \bfy \rangle + c ||\bfy||^2) \bfx + (1 - c ||\bfx||^2) \bfy}{1 + 2 c \langle \bfx, \bfy \rangle + c^2 ||\bfx||^2 ||\bfy||^2}.
\end{equation}
This operation can be used to effectively compute the hyperbolic distance between $\bfx, \bfy \in \mathbb{D}_c^n$ as
\begin{equation}\label{eq:hyp_dist}
    d_c (\bfx, \bfy) = \frac{2}{\sqrt{c}} \tanh^{-1} (\sqrt{c} ||-\bfx \oplus_c \bfy||).
\end{equation}

An important concept in Riemannian geometry is the tangent space $\mathcal{T}_\bfx \mathcal{M}$, which can intuitively be considered as the collection of tangent lines to the manifold $\mathcal{M}$ at the point $\bfx \in \mathcal{M}$. Vectors in this tangent space are called tangent vectors and they represent possible directions and velocities with which one could travel away from the point $\bfx$. Given a tangent vector at a point $\bfx \in \mathcal{M}$, the exponential map at $\bfx$ can be used to determine where one would end up when traveling along the manifold for 1 unit of time in the direction of this tangent vector with a velocity determined by the norm of the vector. For $\mathbb{D}^n$, these exponential maps can be computed as \cite{ganea2018}
\begin{equation}\label{eq:exp_map}
    \exp_\bfx^c (\bfv) = \bfx \oplus_c \Big( \tanh\big(\frac{\sqrt{c} \lambda_\bfx^c ||\bfv||}{2}\big) \frac{\bfv}{\sqrt{c} ||\bfv||} \Big),
\end{equation}
where $\bfv \in \mathcal{T}_\bfx \mathbb{D}_c^n$. The inverses of these exponential maps are the logarithmic maps, given by
\begin{equation}
    \log_{\bfx}^c (\bfy) = \frac{2}{\sqrt{c} \lambda_\bfx^c} \tanh^{-1} (\sqrt{c} ||-\bfx \oplus_c \bfy||) \frac{-\bfx \oplus_c \bfy}{||-\bfx \oplus_c \bfy||},
\end{equation}
where $\bfx, \bfy \in \mathcal{M}$. We can move tangent vectors between tangent spaces by using parallel transport. On the Poincaré ball this parallel transport is given by \cite{shimizu2020}
\begin{equation}
    P_{\bfx \rightarrow \bfy}^c (\bfv) = \frac{\lambda_\bfx^c}{\lambda_\bfy^c} \text{gyr} [\bfy, -\bfx] \bfv,
\end{equation}
where $\text{gyr}[\bfx, \bfy]$ is the gyrator \cite{ungar2022}, defined as
\begin{equation}
    \text{gyr} [\bfx, \bfy] \bfz = -(\bfx \oplus_c \bfy) \oplus_c \big(\bfx \oplus_c (\bfy \oplus_c \bfz)\big),
\end{equation}
for $\bfx, \bfy, \bfz \in \mathbb{D}^n$. All of the mappings introduced here play a crucial role in the definition of hyperbolic networks \cite{ganea2018,shimizu2020}. For notational convenience, the $c$ is dropped from the formulations whenever its value is obvious from the context.

\section{Hyperbolic attacks}
\label{sec:hyperbolic_attacks}
We consider the setting where we are attacking a classification model $\phi: \mathcal{M} \rightarrow \mathbb{R}^C$, with $\mathcal{M}$ an $n$-dimensional Riemannian manifold on which the model input is defined, $C$ the number of output classes, and where $\phi$ is parameterized by parameters $\theta$. Here, $\mathcal{M}$ will usually be either $\mathbb{R}^n$ with the Euclidean distance or $\mathbb{D}^n$ as defined in Section \ref{sec:background}. We note however that our proposed adversarial attacks work for arbitrary $\mathcal{M}$ as long as the exponential map can be computed efficiently. We will use $\bfx \in \mathcal{M}$ and $y \in \{1, \ldots, C\}$ to denote the model input and ground truth label, respectively. The model $\phi$ outputs predictions $\tilde{y}$ as
\begin{equation}
    \tilde{y} = \underset{i \in \{1, \ldots, C\}}{\arg \, \max} \, \phi_i(\bfx).
\end{equation}
Suppose that for some input $\bfx$ the model $\phi$ correctly outputs the true label $\tilde{y} = y$. Then, the objective of an adversarial attack is to generate an adversarial sample $\tilde{\bfx}$ through a slight perturbation of the original input $\bfx$ such that the model $\phi$ will no longer correctly predict the true label $y$ for $\tilde{\bfx}$. Usually the perturbation is constrained in terms of some metric $d: \mathcal{M} \times \mathcal{M} \rightarrow \mathbb{R}$, so
\begin{equation}\label{eq:att_constraint}
    d(\bfx, \tilde{\bfx}) \leq \epsilon,
\end{equation}
where $\epsilon > 0$ is the maximal perturbation size. For example, in the case where $\mathcal{M} = \mathbb{R}^n$, this constraint is typically given in terms of the $L^p$ metrics. Following the notation of \cite{madry18pgd}, the subset of $\mathcal{M}$ that satisfies equation \ref{eq:att_constraint} is denoted by $\mathcal{S}$.

\subsection{Hyperbolic FGM}
The fast gradient method (FGM) is an efficient attack defined for Euclidean inputs which generates adversarial samples through a single update \cite{goodfellow15fgsm}. Given some input $\bfx \in \mathbb{R}^n$, an adversarial sample is obtained through computing
\begin{equation}\label{eq:fgm_original_update}
    \tilde{\bfx} = \bfx + \alpha \nabla_\bfx J(\mathbf{\theta}, \bfx, y),
\end{equation}
where $J$ is some objective function (typically the same as the one used for training the model $\phi$) and where $\alpha > 0$ is the magnitude of the attack. The value of $\alpha$ can be chosen such that $\tilde{\bfx} \in \mathcal{S}$.
This process is essentially just typical gradient ascent on some objective function that is assumed to be inversely related to the model's performance. Therefore, inspired by \cite{bonnabel2013}, this process can be generalized to an arbitrary Riemannian manifold as long as the exponential maps can be computed efficiently. More precisely, we can define Riemannian FGM with $\bfx \in \mathcal{M}$ as
\begin{equation}
    \tilde{\bfx} = \exp_{\bfx} \big(\alpha \nabla_\bfx J(\mathbf{\theta}, \bfx, y)\big),
\end{equation}
where $\exp_\bfx: \mathcal{T}_\bfx \mathcal{M} \rightarrow \mathcal{M}$ is the exponential map at $\bfx$. Note that this method is indeed equivalent to the original FGM when applied to Euclidean space, since the exponential maps for Euclidean space can be defined as
\begin{equation}
    \exp_\bfx (\mathbf{v}) = \bfx + \mathbf{v}.
\end{equation}
Hyperbolic FGM on the Poincaré ball can then be defined as a special case of Riemannian FGM, where the exponential maps are defined as in Equation \ref{eq:exp_map}.

It should be mentioned here that a commonly used attack is the fast gradient sign method (FGSM), which uses the sign of the gradient instead of the gradient itself in equation \ref{eq:fgm_original_update}. This method is often applied in the setting where the constraint in equation \ref{eq:att_constraint} is given in terms of the $L^\infty$ metric. Under this constraint, the FGM and FGSM updates usually result in identical adversarial samples. When using for example the $L^2$ norm, it makes more sense to use the full information given by the gradient, which is why we have chosen to use FGM instead of FGSM.

\subsection{Hyperbolic PGD}
The projected gradient descent (PGD) attack is essentially a multi-step version of FGM where, after each step, the output is projected back onto $\mathcal{S}$ \cite{kurakin17pgd,madry18pgd}. More specifically, each PGD update is computed as
\begin{equation}\label{eq:pgd_update}
    \tilde{\bfx}^{t + 1} = \pi_{\mathcal{S}} \Big(\tilde{\bfx}^t + \alpha \nabla_\bfx J(\mathbf{\theta}, \tilde{\bfx}^t, y)\Big),
\end{equation}
where $\tilde{\bfx}^0 = \bfx \in \mathbb{R}^n$, where $\alpha > 0$ is the step size and where $\pi_{\mathcal{S}}: \mathbb{R}^n \rightarrow \mathcal{S}$ is some projection from $\mathbb{R}^n$ onto $\mathcal{S}$. This projection onto $\mathcal{S}$ is usually taken to be the shortest path projection, so
\begin{equation}
    \pi_{\mathcal{S}} (\bfy) = \underset{\bfz \in \mathcal{S}}{\arg \min} \; d(\bfy, \bfz),
\end{equation}
for $\bfy \in \mathbb{R}^n$. The number of updates that is performed is denoted by $T$ and the resulting adversarial sample by $\tilde{\bfx}^T$.

Similarly, we can define Riemannian PGD from Riemannian FGM by simply performing multiple steps of Riemannian FGM with projections in between. In other words, for an arbitrary Riemannian manifold $\mathcal{M}$ we can define Riemannian PGD as
\begin{equation}
    \tilde{\bfx}^{t + 1} = \pi_{\mathcal{S}} \Big(\exp_{\tilde{\bfx}^t} \big(\alpha \nabla_\bfx J(\mathbf{\theta}, \tilde{\bfx}^t, y)\big)\Big),
\end{equation}
where $\pi_\mathcal{S}: \mathcal{M} \rightarrow \mathcal{S}$ is a projection from $\mathcal{M}$ onto $\mathcal{S}$. When applied to Euclidean space this equation simplifies to the original PGD update from equation \ref{eq:pgd_update}. Hyperbolic PGD is again a specific case of Riemannian PGD, where we simply apply the exponential map as defined in equation \ref{eq:exp_map} and where $\pi_\mathcal{S}$ is the shortest path projection with respect to the hyperbolic distance.

\subsection{Objective functions}
There are several interesting objective functions that can be considered for computing the gradients used by the attacks. The most commonly used \cite{goodfellow15fgsm,kurakin17pgd,madry18pgd} objective function is the usual cross entropy loss, given by
\begin{equation}
    J_{\text{CE}} (\theta, \bfx, y) = -\log \frac{e^{\phi_y (\bfx)}}{\sum_{i=1}^C e^{\phi_i (\bfx)}}.
\end{equation}
As this loss has been used to train the model, perturbing the input to increase this loss can be expected to efficiently degrade model performance. 

Using cross entropy loss, however, may not always lead to optimal attacks, as demonstrated for instance by \cite{croce2020autoattack}. Under the assumption that the model is correct for the original input, there are several other interesting options where the perturbations are aimed at directly changing specific logits. For example, we can use the negative value of the maximal logit 
\begin{equation}
    J_{\text{NFL}} = - \max_i \phi_i (\bfx),
\end{equation}
which will push the model away from the correct label. Alternatively, we can target the second largest logit
\begin{equation}
    J_{\text{SL}} = \max_{i \neq \arg \max_{j} \phi_j (\bfx)} \phi_i (\bfx),
\end{equation}
which steers the model towards the second most probable class. Lastly, the smallest logit can be used as the objective function
\begin{equation}
    J_{\text{LL}} = \min_i \phi_i (\bfx).
\end{equation}
This pushes the model towards predicting the least likely class.

\section{Synthetic example}
\label{sec:toy_example}

To gain some insights into the behaviour of the hyperbolic attacks compared to the original attacks we perform a small synthetic classification experiment with data on the Poincaré disk with curvature $-1$, so where $c = 1$.

\subsection{Dataset}
\begin{figure}[t]
    \centering
    \begin{subfigure}[t]{0.45\textwidth}
        \includegraphics[width=\textwidth]{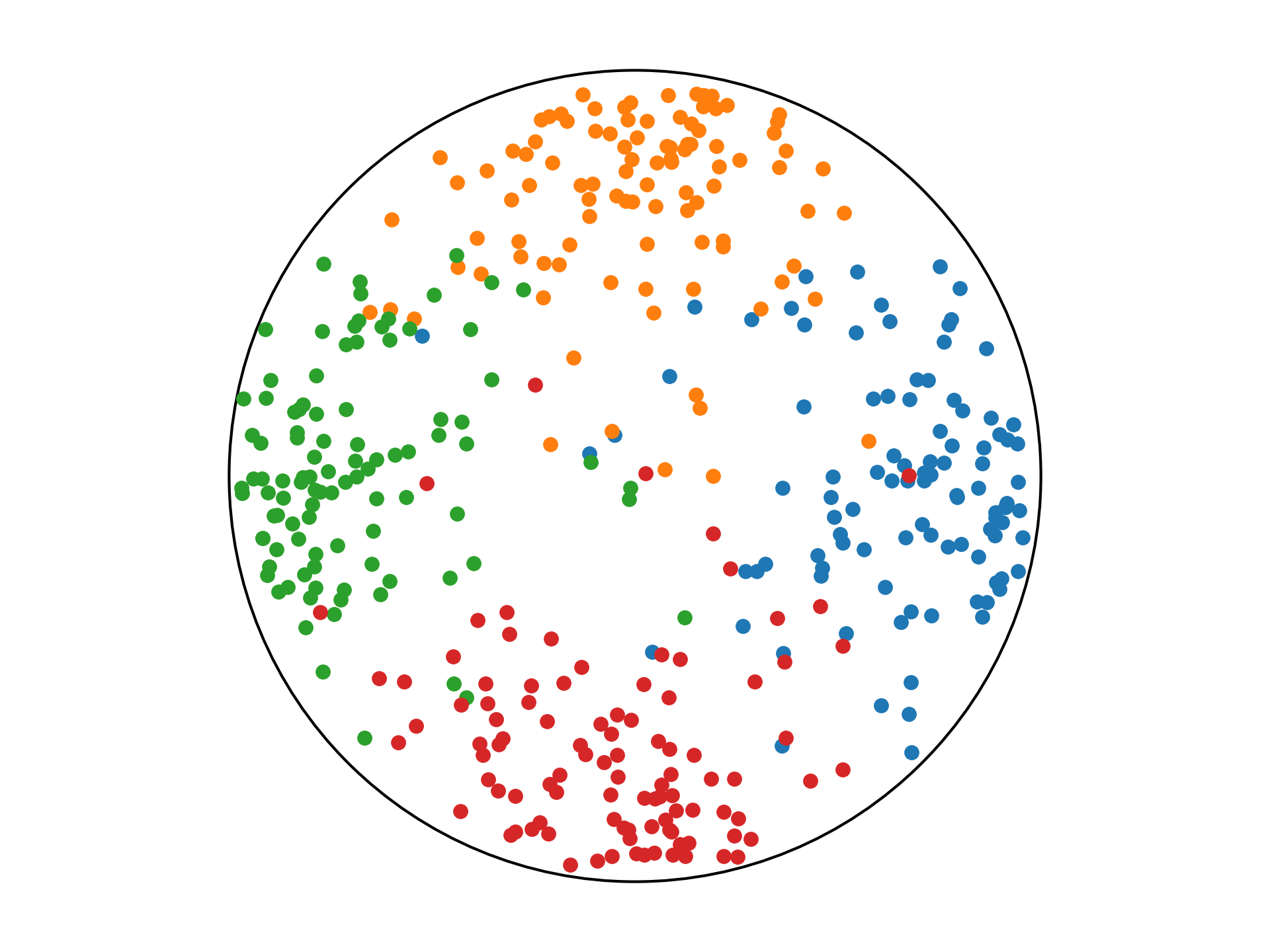}
        \caption{Example of 100 generated samples for each of the $4$ classes with the $\mu_i$ evenly spaced on a circle centered at the origin with a radius of $r = 1.5$ with respect to the hyperbolic distance and with a variance of $\sigma^2 = 0.25$.}
        \label{fig:dataset}
    \end{subfigure}
    \hspace{0.03\textwidth}
    \begin{subfigure}[t]{0.50\textwidth}
        \includegraphics[width=\textwidth]{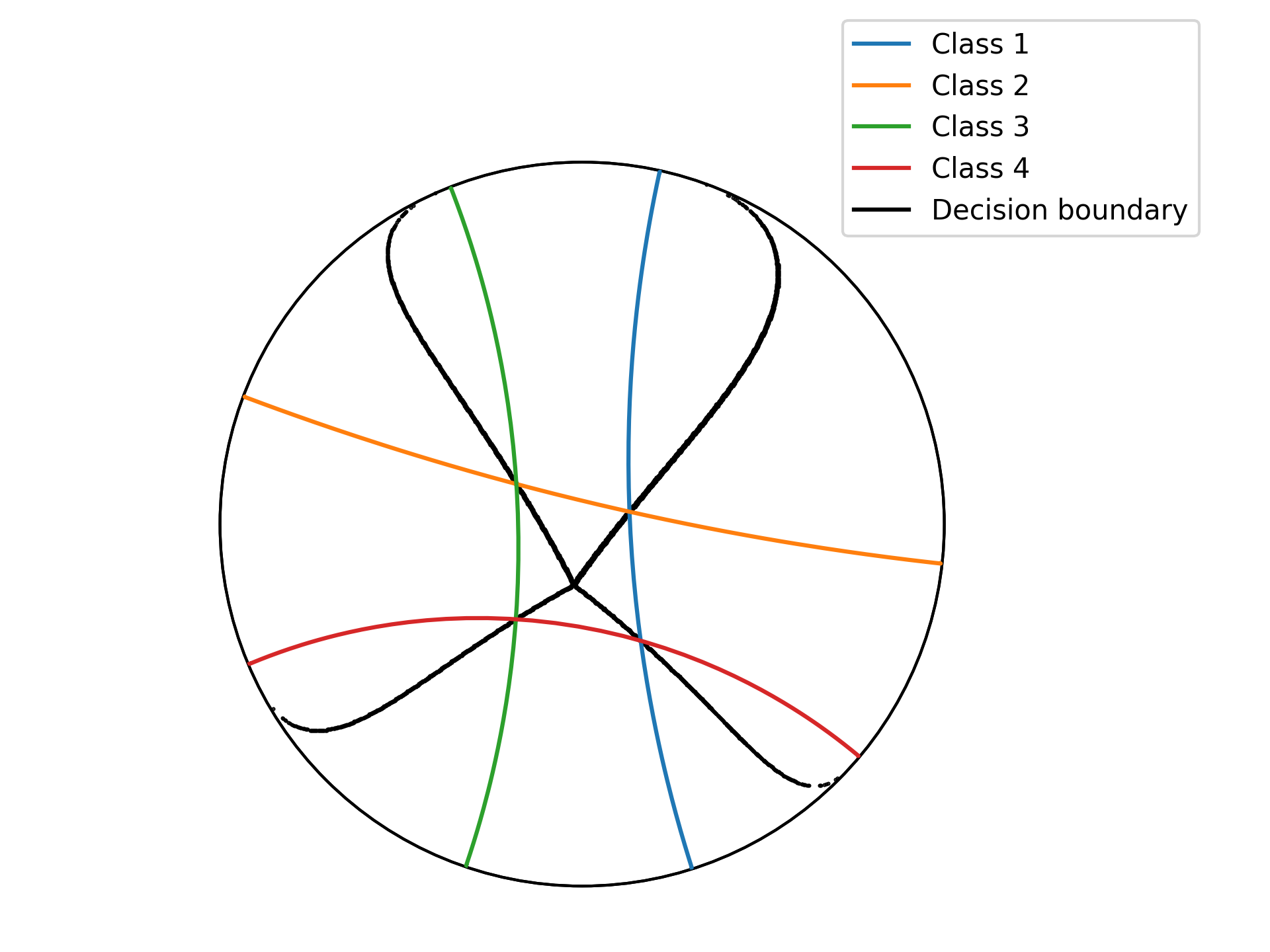}
        \caption{Hyperplanes and decision boundary of the simple hyperbolic MLR model trained on data generated from the same distribution as shown in Figure \ref{fig:dataset}.}
        \label{fig:mlr_model}
    \end{subfigure}
    \vspace{-5pt}
    \caption{Visualizations of the generated dataset and the hyperbolic MLR model.}
    \label{fig:enter-label}
    \vspace{-10pt}
\end{figure}

We generate data by sampling from 4 different wrapped normal distributions \cite{nagano2019} with covariance matrices $\Sigma_i \in \mathbb{R}^{2 \times 2}$ and means $\mu_i \in \mathbb{D}^2$ for $i \in \{1, \ldots, 4\}$. Sampling from such a distribution works as follows:
\begin{enumerate}
    \item Sample points $\hat{\bfx}_j$ from a multivariate normal distribution $\mathcal{N}(\mathbf{0}, \Sigma_i)$ for $j = 1, \ldots, N_i$, where $N_i$ is the number of points for the $i$-th class.
    \item Interpret the sampled points as tangent vectors at the origin of the disk, so $\hat{\bfx}_j \in \mathcal{T}_\mathbf{0} \mathbb{D}^2$.
    \item Use parallel transport to transport these tangent vectors to the tangent space at $\mu_i$, so compute $P_{\mathbf{0} \rightarrow \mu_i} (\hat{\bfx}_j) \in \mathcal{T}_{\mu_i} \mathbb{D}^2$.
    \item Map the tangent vectors to the disk using the exponential map at $\mu_i$, i.e. $x_j = \exp_{\mu_i} (P_{\mathbf{0} \rightarrow \mu_i} (\hat{\bfx}_j))$.
\end{enumerate}
In our experiments we choose the $\mu_i$ to be evenly spaced points along a circle in the Poincaré disk with the origin as its center and with radius $r = 1.5$ with respect to the hyperbolic distance. We set each covariance matrix to $\Sigma_i = \sigma^2 I_2$, where $I_2$ is the 2-dimensional identity matrix and with $\sigma^2 = 0.25$. The generated training dataset and test dataset each contain 10K samples for each class. Some example data generated with this setup is shown in Figure \ref{fig:dataset}.

\subsection{Model}
We train a simple hyperbolic multinomial logistic regression model \cite{ganea2018,shimizu2020} to perform the classification. The parameters of this model define $4$ hyperplanes, one for each class, and the output logits are computed as the signed distances of the input to each of these hyperplanes. For the training setup we use Riemannian SGD \cite{bonnabel2013} with a learning rate of $5 * 10^{-5}$ and a batch size of 4096 for 100 epochs. The model achieves an accuracy of 86.7\% on the test data. Figure \ref{fig:mlr_model} shows the learned hyperplanes and an approximation of the decision boundaries.

\vspace{-8pt}

\subsection{Attack configuration}
\vspace{-5pt}
We attack the resulting model with the proposed hyperbolic FGM. The constraint on the perturbation size in equation \ref{eq:att_constraint} is defined in terms of the hyperbolic distance as defined in equation \ref{eq:hyp_dist}. The step size is set to
\begin{equation}
    \alpha = \frac{\epsilon}{\lambda_\bfx^c ||\nabla_\bfx J(\theta, \bfx, y)||},
\end{equation}
such that $d_c (\bfx, \tilde{\bfx}) = \epsilon$. For the objective function we try each of the options discussed in Section \ref{sec:hyperbolic_attacks} to see their effects on the resulting attack. 

For comparison, the model is attacked naively with the original FGM as well. Note that this attack ignores the geometry of the manifold, which makes it particularly inappropriate for this kind of data. Moreover, the choice of projection $\pi_\mathcal{S}$ is less obvious in this case since the perturbation is not along a geodesic, meaning that orthogonal projection onto $\mathcal{S}$ is generally not equivalent to picking a suitable step size. To stay close to the true nature of this attack, we opt to pick $\alpha$ such that $d_c (\bfx, \tilde{\bfx}) = \epsilon$ here as well, which is equivalent to projection along a Euclidean straight line. We use Newton's method to find the value of $\alpha$ that satisfies this condition.

\vspace{-8pt}

\subsection{Results}
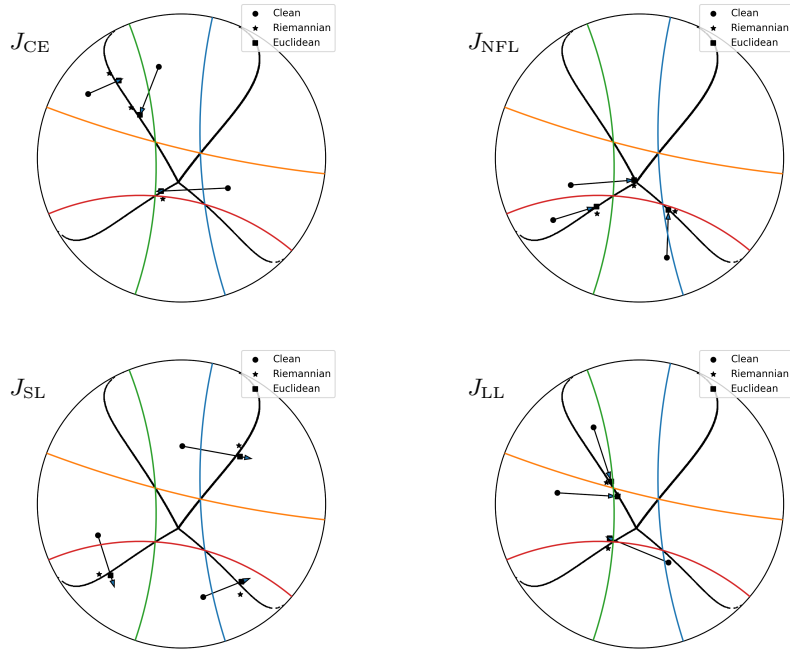
\begin{figure}
    \centering
    \begin{subfigure}{0.49\textwidth}
        \begin{picture}(100, 130)
            \put(0,0){\includegraphics[width=\textwidth]{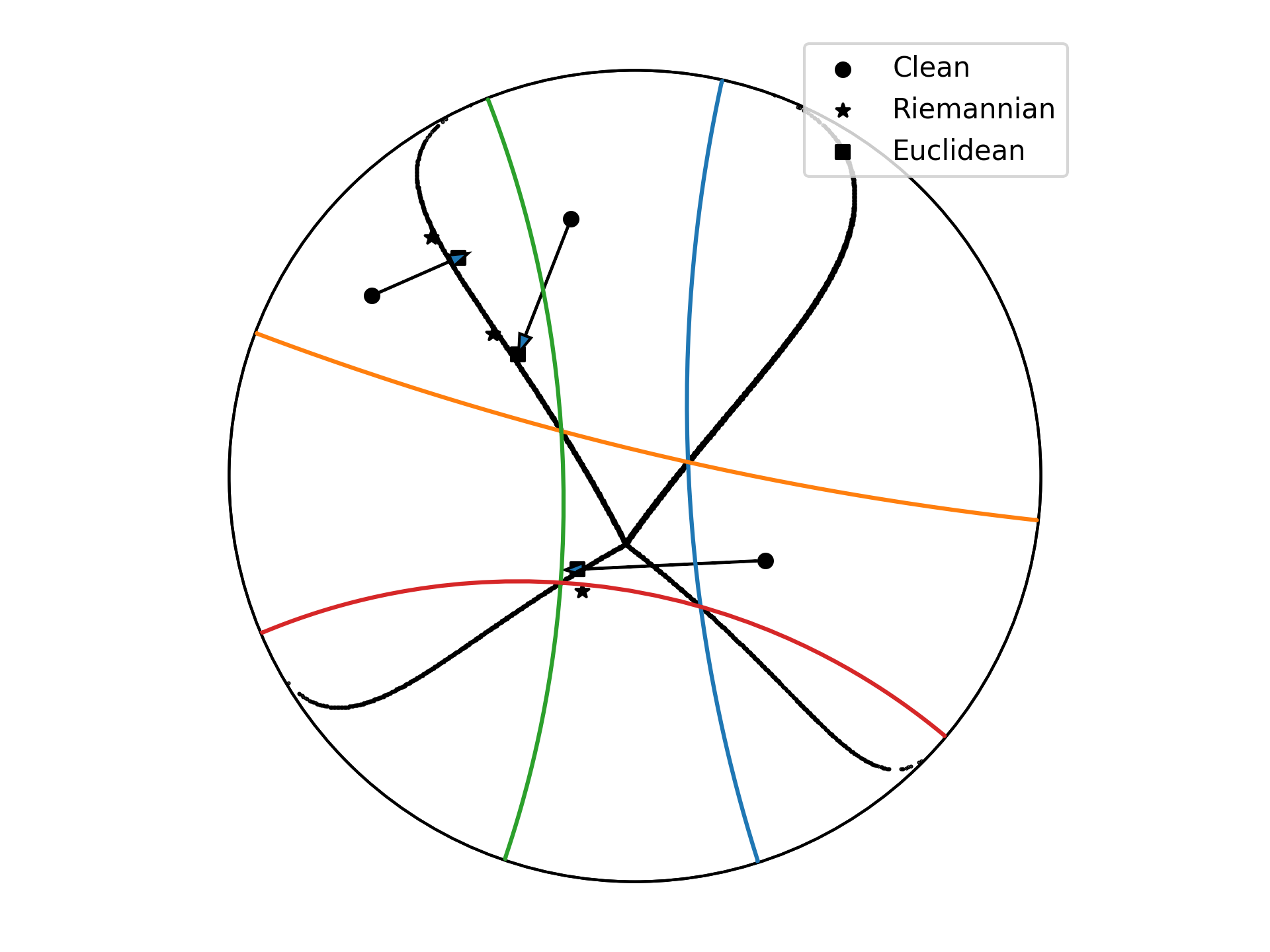}}
            \put(20, 105){$J_{\text{CE}}$}
        \end{picture}
    \end{subfigure}
    \begin{subfigure}{0.49\textwidth}
        \begin{picture}(100, 130)
            \put(0,0){\includegraphics[width=\textwidth]{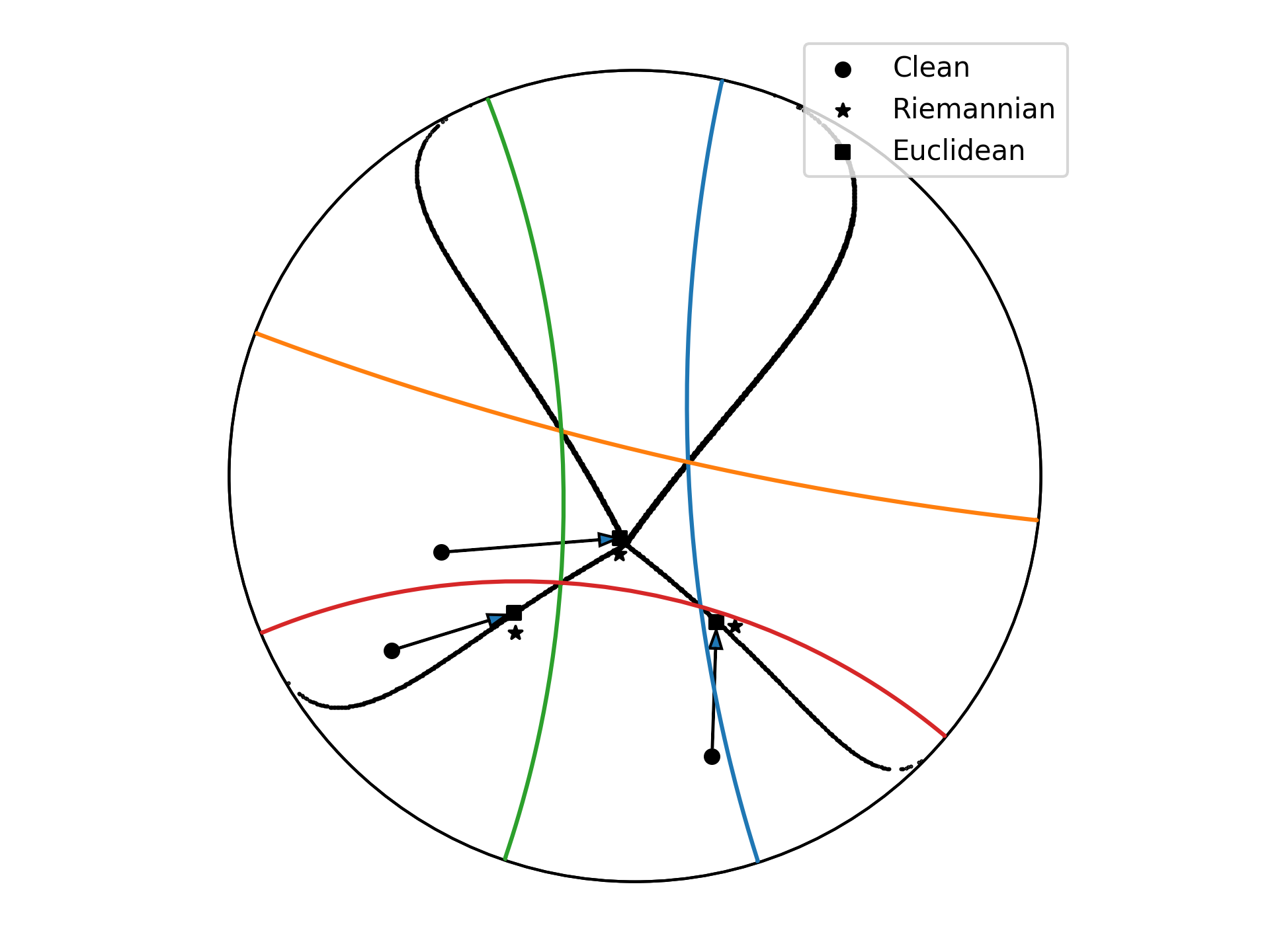}}
            \put(20, 105){$J_{\text{NFL}}$}
        \end{picture}
    \end{subfigure}
    \begin{subfigure}{0.49\textwidth}
        \begin{picture}(100, 130)
            \put(0,0){\includegraphics[width=\textwidth]{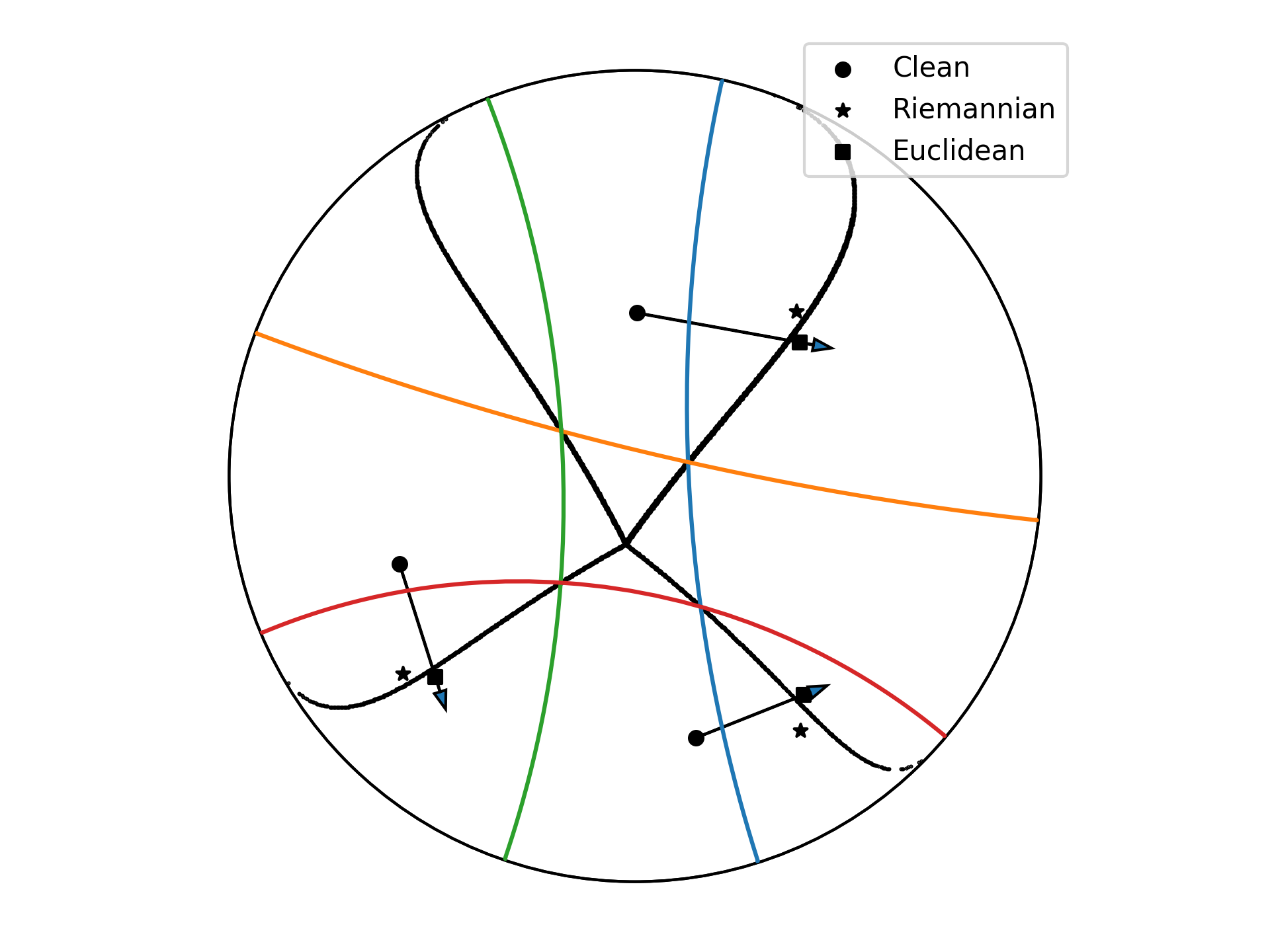}}
            \put(20, 105){$J_{\text{SL}}$}
        \end{picture}
    \end{subfigure}
    \begin{subfigure}{0.49\textwidth}
        \begin{picture}(100, 130)
            \put(0,0){\includegraphics[width=\textwidth]{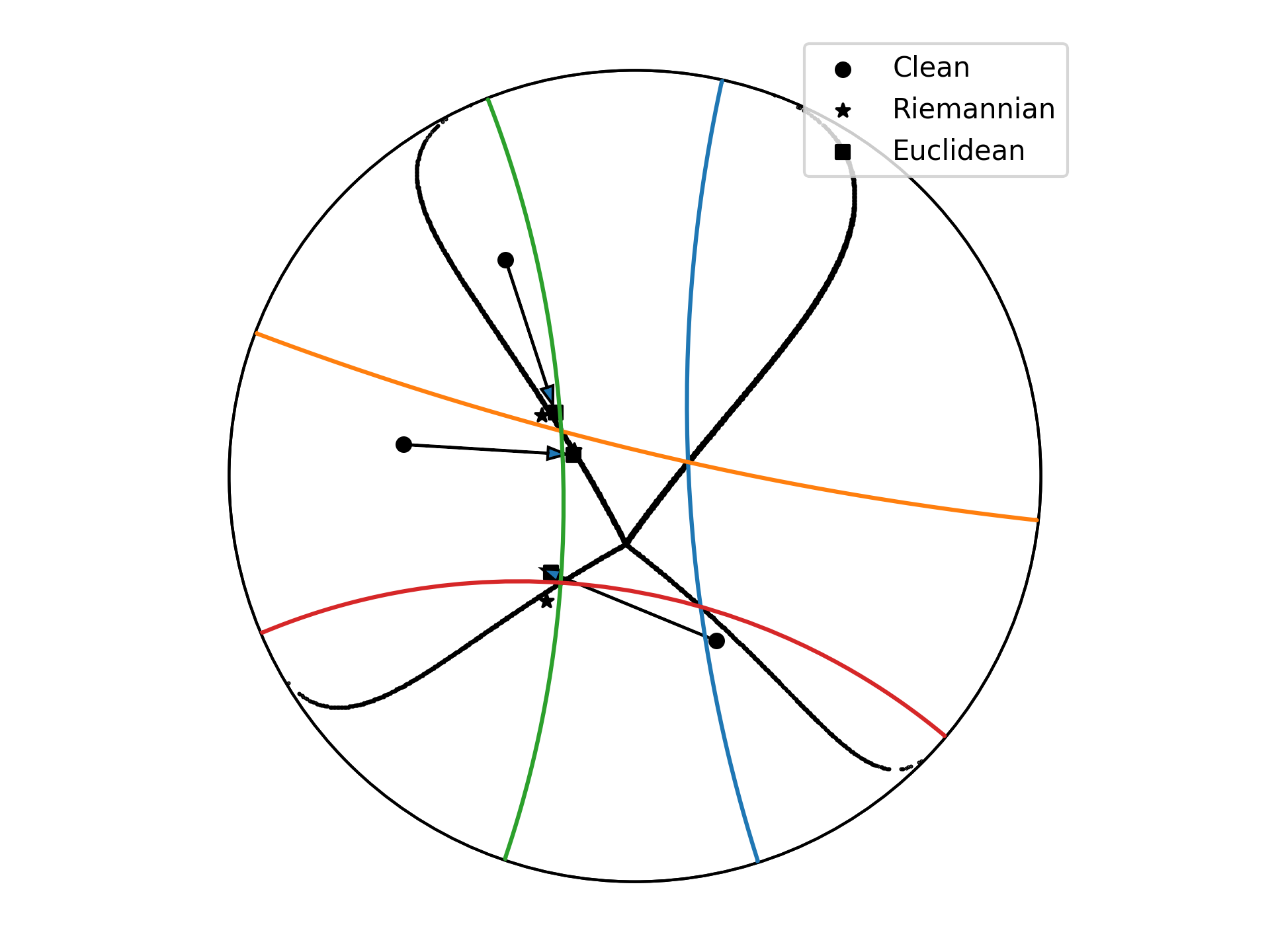}}
            \put(20, 105){$J_{\text{LL}}$}
        \end{picture}
    \end{subfigure}
    \caption{Examples of adversarial samples generated with the hyperbolic (Riemannian) FGM or the original Euclidean FGM with the four different objective functions and a perturbation size $\epsilon = 1.0$. The arrows represent the gradient used for the perturbation. When this gradient points inward towards the origin, the hyperbolic attack seems stronger, while the original FGM appears to be more powerful when the gradient points outwards.}
    \label{fig:toy_example}
\end{figure}

\vspace{-5pt}
Figure \ref{fig:toy_example} shows several example adversarial samples that were generated with the objective functions from Section \ref{sec:hyperbolic_attacks} and with either Riemannian FGM or, naively, with the original FGM. For each of the samples shown, the model predicts a different label for the two adversarial samples generated with either attack. When using $J_{\text{CE}}$ we find that the gradient generally points somewhat in the direction of the hyperplane corresponding to the current prediction, while still being quite sensitive to the sample's position with respect to the other hyperplanes. On the other hand, when using $J_{\text{NFL}}$ the gradient points nearly exactly in the direction of the geodesic that forms the shortest path to the hyperplane of the correct class. Similarly, $J_{\text{SL}}$ and $J_{\text{LL}}$ lead to gradients that point in the direction of the geodesic forming the shortest path to the hyperplanes corresponding to the targeted logits.

\begin{figure}
    \centering
    \begin{subfigure}{0.49\textwidth}
        \includegraphics[width=\textwidth]{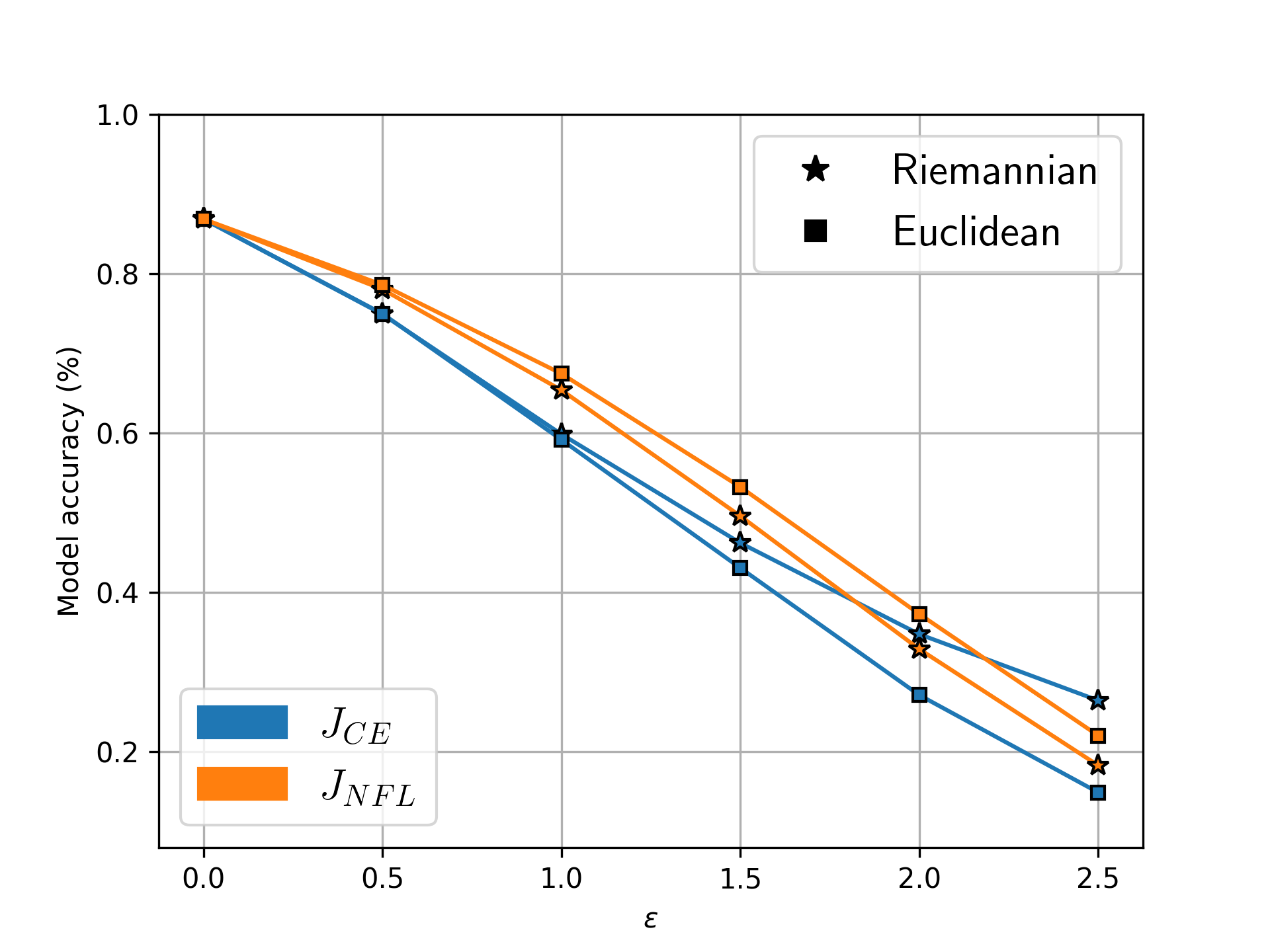}
        \caption{$J_{\text{CE}}$ and $J_{\text{NFL}}$}
    \end{subfigure}
    \begin{subfigure}{0.49\textwidth}
        \includegraphics[width=\textwidth]{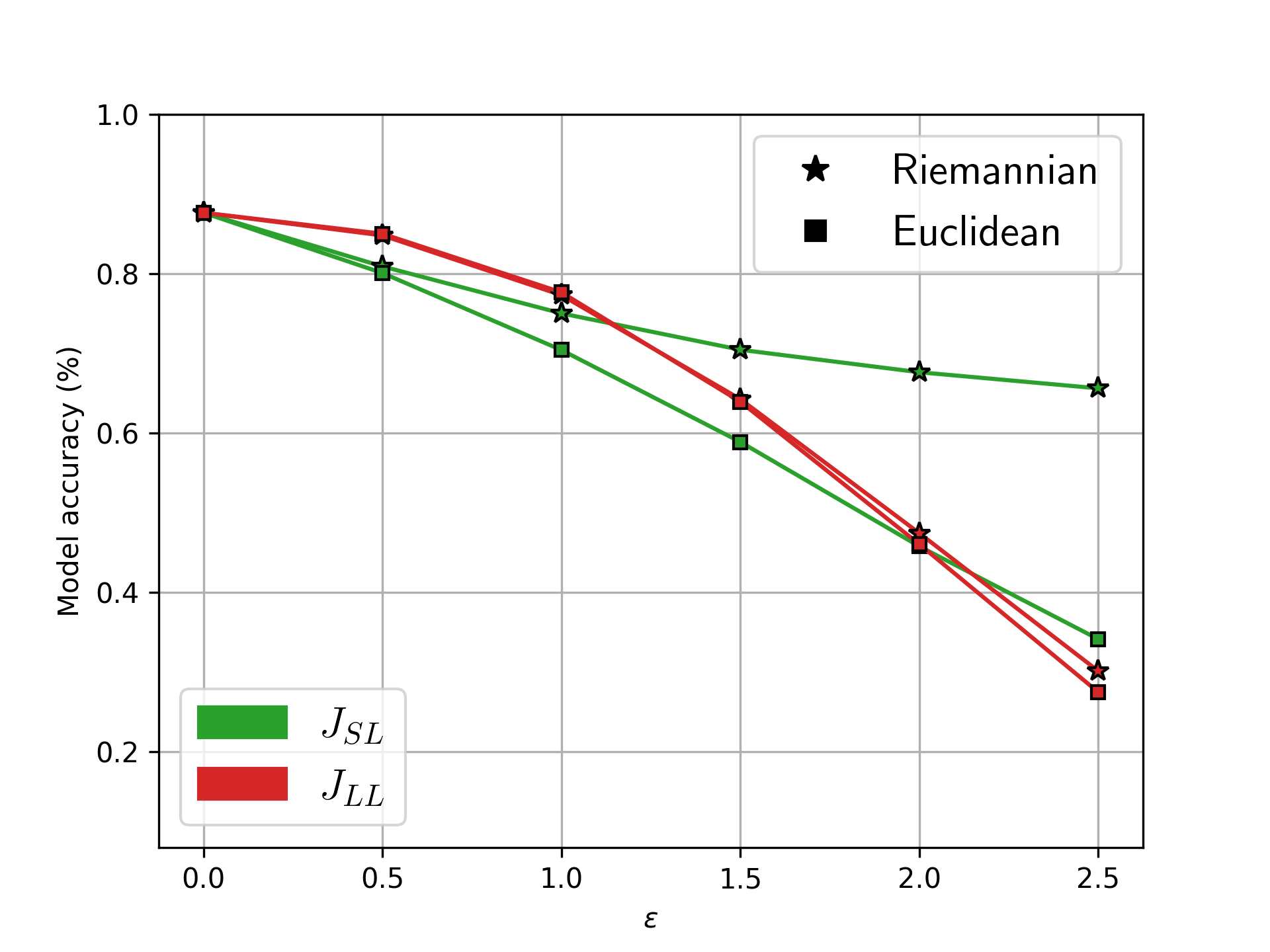}
        \caption{$J_{\text{SL}}$ and $J_{\text{LL}}$}
    \end{subfigure}
    \vspace{-8pt}
    \caption{The performance of the hyperbolic MLR model when attempting to classify adversarial samples generated with the different attacks for varying values of $\epsilon$.}
    \label{fig:toy_results}
    \vspace{-20pt}
\end{figure}

Futhermore, we oberserve that for gradients pointing inward towards the origin, the hyperbolic attack seems to be more potent, while for other gradients the Euclidean attack seems stronger. As the hyperbolic attack generates adversarial samples along geodesics of the Poincaré ball, samples generated with this attack will curve outwards towards the boundary as $\epsilon$ increases. This seems to be beneficial when the gradient points inward and the corresponding geodesic curves outward slowly, but detrimental when the gradient points away from the origin. The accuracy of the model after the various attacks with varying $\epsilon$ is shown in Figure \ref{fig:toy_results}. From these results it appears that the best objective function is $J_{\text{CE}}$ and that the Euclidean attack performs better for this objective. The second best objective is $J_{\text{NFL}}$ for which the hyperbolic attack outperforms the Euclidean attack. However, it should be noted that these results are highly sensitive to the configuration of the dataset and the model parameters.

\vspace{-20pt}
\vspace{-10pt}

\section{Hyperbolic models}
\label{sec:hyperbolic_models}
\vspace{-5pt}
In this section, we will take a look at the differences in adversarial robustness between Poincaré \cite{spengler2023} and Euclidean ResNets \cite{he2016}. Moreover, we will apply the proposed hyperbolic attacks to the Poincaré ResNets to see if accounting for the geometry resolves these differences in adversarial performance.
A Poincaré ResNet is a fully hyperbolic version of the original ResNet, where each of the layers appearing in the network is replaced by a fully hyperbolic alternative, based on \cite{ganea2018,shimizu2020}. To apply these networks to images, the image is first mapped to hyperbolic space by considering each pixel vector to be a tangent vector at the origin of $\mathbb{D}^3$ and using the exponential map corresponding to this tangent space. This operation can be seen as a preprocessing step. The resulting network has been found to be more adversarially robust than its Euclidean counterpart \cite{spengler2023}.

\vspace{-8pt}

\subsection{Datasets and models}
\vspace{-5pt}
The experiments in this section are performed on the CIFAR-10 and CIFAR-100 datasets. On each dataset, we train a Poincaré ResNet \cite{spengler2023} and a Euclidean ResNet \cite{he2016} each with a depth of 20 or 32 layers and channel widths 8, 16 and 32. For the Poincaré ResNets we use the Riemannian Adam \cite{becigneul2018} optimizer with a learning rate of $1 * 10^{-3}$ and a weight-decay of $1 * 10^{-4}$. For the Euclidean ResNets we use the Adam optimizer with the same learning rate and weight-decay. In each setting we train for 500 epochs. For the Poincaré ResNets the curvature is fixed to a value of $-0.1$, so where $c = 0.1$. Lastly, for data augmentation we apply random cropping and horizontal flipping. The accuracies that these models achieve on the test splits are shown in Table \ref{tab:model_accuracies}.

\vspace{-8pt}

\begingroup
\setlength{\tabcolsep}{10pt}
\begin{table}[t]
    \centering
    \begin{tabular}{c cc cc}
        \toprule
         & \multicolumn{2}{c}{CIFAR-100} & \multicolumn{2}{c}{CIFAR-10} \\
         \cmidrule(lr){2-3} \cmidrule(lr){4-5}
         & 20 & 32 & 20 & 32 \\
         \midrule
        Poincaré ResNet & 54.1 & 57.5 & 86.2 & 87.5 \\
        Euclidean ResNet & 54.3 & 56.4 & 84.9 & 86.0 \\
        \bottomrule
    \end{tabular}
    \caption{Accuracy (\%) for the Poincaré and Euclidean ResNets with depths of either 20 or 32 layers and channel widths 8, 16 and 32 on the test splits of CIFAR-10 and CIFAR-100. Hyperbolic and Euclidean ResNets obtain similar classification performance.}
    \label{tab:model_accuracies}
    \vspace{-15pt}
\end{table}
\endgroup

\subsection{Poincaré ResNets versus Euclidean ResNets}
\begin{figure}[t]
    \centering
    \begin{subfigure}[t]{0.49\textwidth}
        \includegraphics[width=\textwidth]{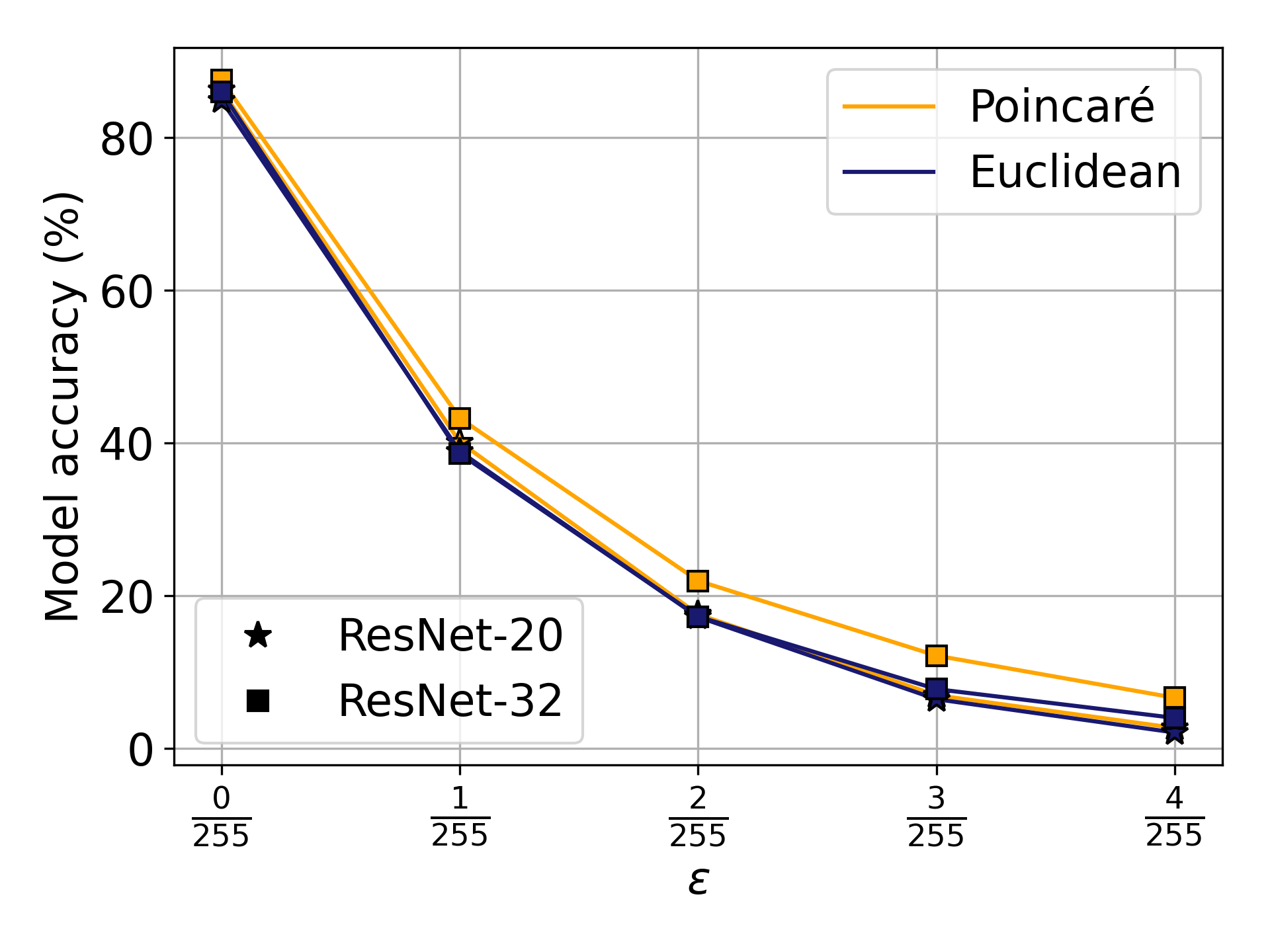}
        \caption{Original FGSM}
        \label{fig:fgsm_attacked_model_performance}
    \end{subfigure}
    \begin{subfigure}[t]{0.49\textwidth}
        \includegraphics[width=\textwidth]{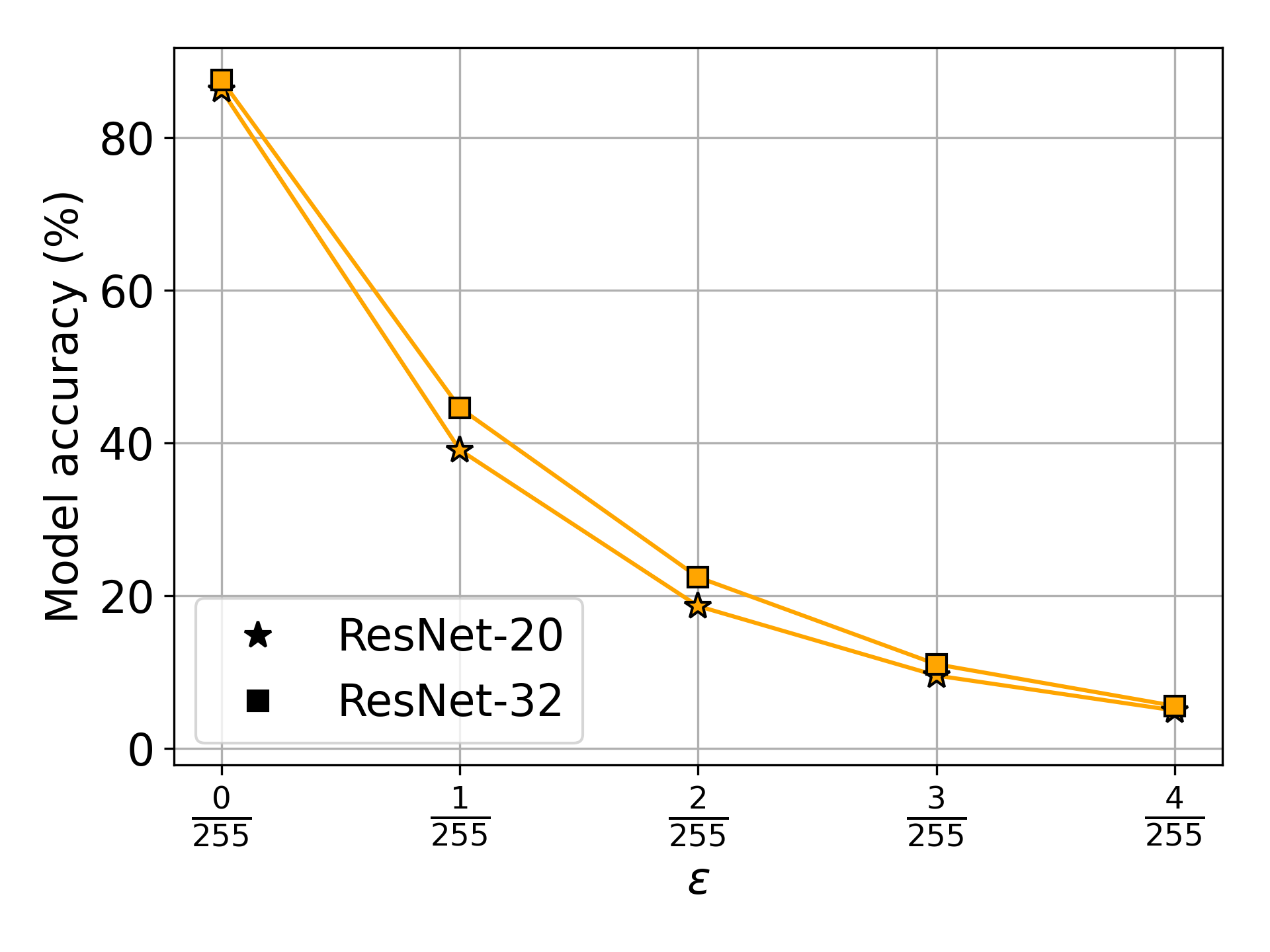}
        \caption{Hyperbolic FGSM}
        \label{fig:rfgsm_attacked_model_performance}
    \end{subfigure}
    \caption{Accuracy on CIFAR-10 of the Poincaré and Euclidean ResNets with depths 20 or 32 when attacked using the original FGSM or the hyperbolic FGSM with increasingly large perturbation size $\epsilon$.}
    \label{fig:attacked_model_performance}
    \vspace{-20pt}
\end{figure}

\vspace{-5pt}
We attack each of the ResNet models with the original FGM and PGD attacks. Note that this is possible for Poincaré ResNets since we can backpropagate through the exponential map that is applied as a preprocessing step. The results of this attack for FGM on CIFAR-10 are shown in Figure \ref{fig:fgsm_attacked_model_performance}. The results for CIFAR-100 show a very similar pattern, so these have been left out of here for brevity. For PGD we set the step size equal to $0.5 \epsilon$ and $T = 10$. We found that this gives nearly identical results as FGM, so for clarity we will only show the results of FGM. The results for FGM show that the Poincaré and Euclidean ResNet-20 perform similarly against the attack. However, the Poincaré ResNet-32 appears notably more robust than its Euclidean counterpart. 

\begin{figure}[t]
    \centering
    \begin{subfigure}[t]{0.45\textwidth}
        \includegraphics[width=\textwidth]{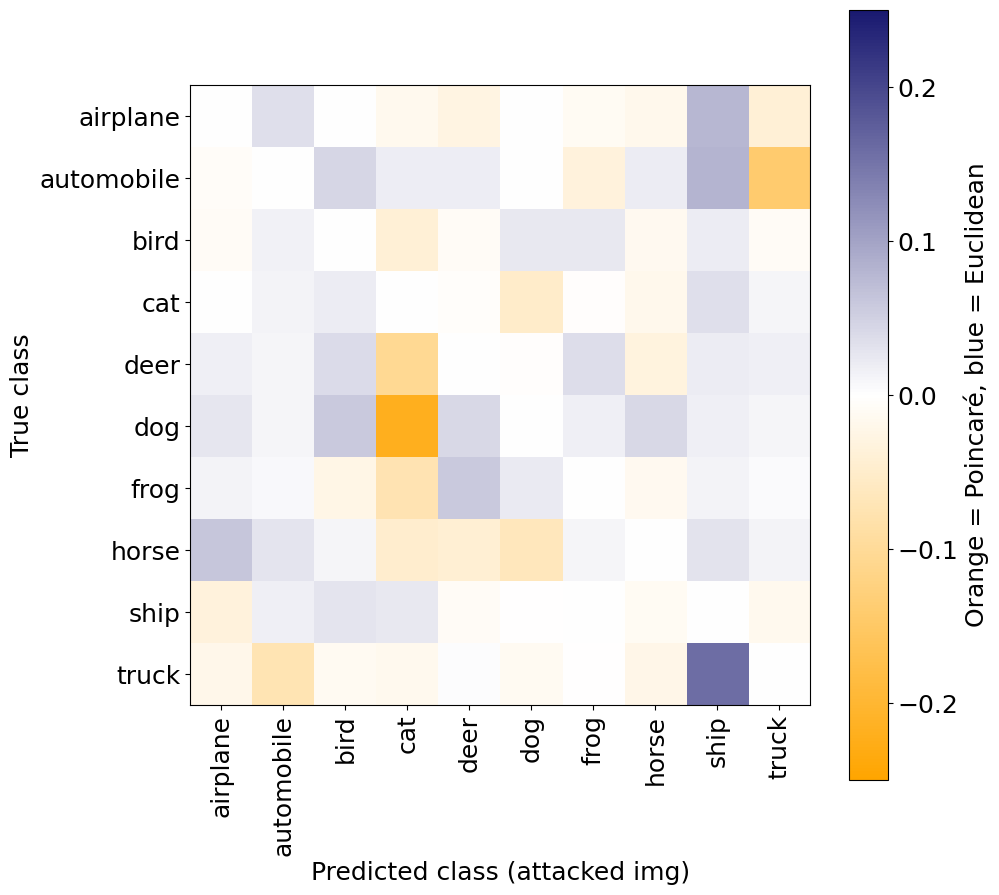}
        \caption{Original FGSM on Poincaré ResNets vs. Euclidean ResNets.}
        \label{fig:misclass_mat_fgsm}
    \end{subfigure}
    \hspace{0.08\textwidth}
    \begin{subfigure}[t]{0.45\textwidth}
        \includegraphics[width=\textwidth]{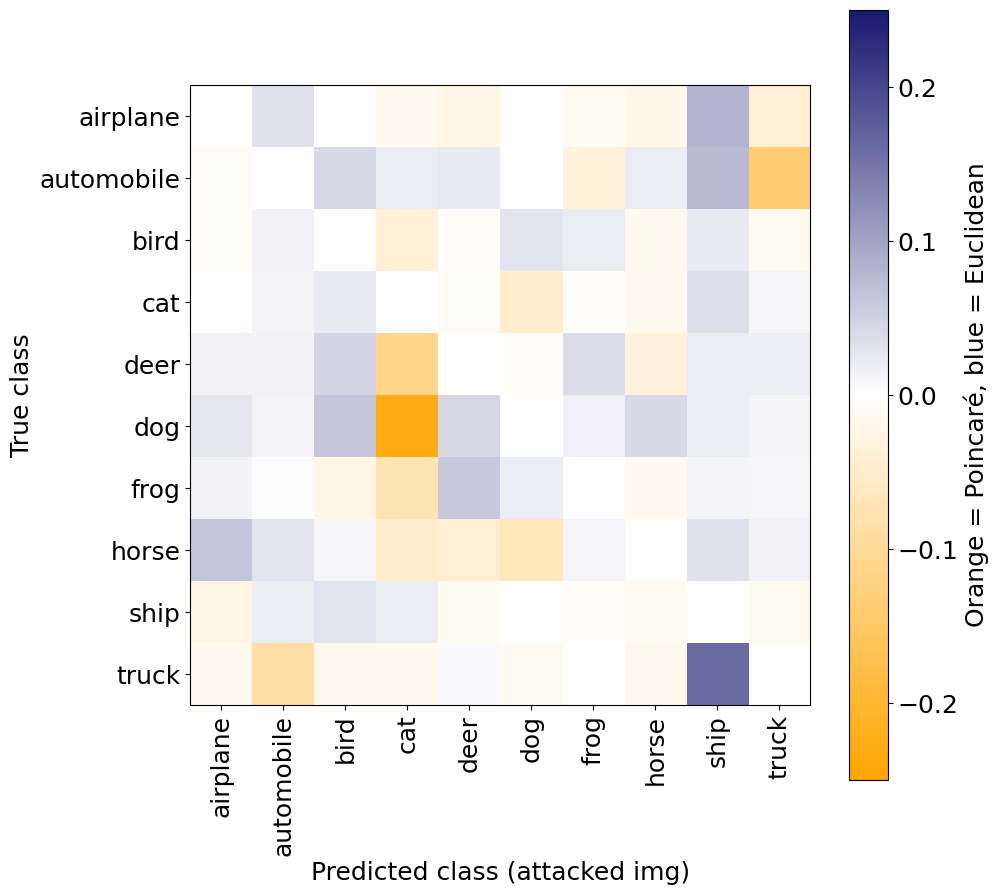}
        \caption{Hyperbolic FGSM on Poincaré ResNets vs. original FGSM on Euclidean ResNets.}
        \label{fig:rfgsm_vs_fgsm}
    \end{subfigure}
    \caption{Comparative misclassification matrix $M^{comp}$ showing for each label $i$, determined by the row, which of the models are easier to trick into predicting label $j$, determined by the column, when applying the original FGSM attack. Orange indicates that the hyperbolic models are easier to trick into the corresponding conversion, while blue indicates a weakness for the Euclidean models.}
    \label{fig:misclass_mats}
    \vspace{-18pt}
\end{figure}

For additional insights into the weaknesses of these models, we visualize the comparative misclassification matrix $M^{comp}$ for CIFAR-10 in Figure \ref{fig:misclass_mat_fgsm}. The value at position $(i, j)$ in this matrix represents the difference between the Poincaré ResNet and the Euclidean ResNet in proportion of misclassified images with true label determined by the row number $i$ that was misclassified as the label determined by $j$, so
\vspace{-2pt}
\begin{equation}
    M^{comp} = M^{P} - M^{E},
\end{equation}
where $M^P$ and $M^E$ are the relative misclasification matrices for the Poincaré ResNets and Euclidean ResNets, respectively, defined as
\begin{equation}
    M_{ij}^{P} = \frac{|\{(\bfx_k, y_k) : y_k = i, \phi^P (\bfx_k) = j\}|}{|\{(\bfx_k, y_k) : y_k = i, \phi^P(\bfx_k) \neq y_k\}|},
\end{equation}
and
\begin{equation}
    M_{ij}^{E} = \frac{|\{(\bfx_k, y_k) : y_k = i, \phi^P (\bfx_k) = j\}|}{|\{(\bfx_k, y_k) : y_k = i, \phi^P(\bfx_k) \neq y_k\}|},
\end{equation}
where $\phi^P$ and $\phi^E$ are the Poincaré and Euclidean ResNets, respectively, and where $i$ and $j$ are the labels corresponding to the $i$-th row and $j$-th column of these matrices. To get a single matrix, the matrices $M^P$ and $M^E$ are averaged over the different Poincaré and Euclidean ResNets, respectively, and over the different values of $\epsilon$. Intuitively, the matrix $M^{comp}$ shows where these models' weaknesses differ from each other by highlighting where models are tricked relatively often into some wrong prediction.

As clearly seen from Figure \ref{fig:misclass_mat_fgsm}, the matrix contains quite a few values that are far from 0. For example, the hyperbolic model is easily tricked into mistaking a dog for a cat, while the Euclidean model easily mistakes a truck for a ship when the images are attacked. While a bit harder to visualize due to the number of classes, a similar pattern appears for CIFAR-100, which is again left out for brevity. These results demonstrate that the models indeed have significantly different weaknesses. So, the choice of geometry somehow leads to different behaviour when facing adversarial attacks.

\vspace{-8pt}

\subsection{Hyperbolic attacks versus Euclidean attacks}
\vspace{-5pt}
One possible explanation for the difference in behaviour between these models is the fact that we attacked the hyperbolic model without accounting for the geometry. To see if this is the case we compare the behaviour of the Poincaré ResNets when attacked with the proposed hyperbolic FGM versus the behaviour of the Euclidean ResNets attacked with the original FGM. In order to apply the hyperbolic FGM, we have to apply it to the preprocessed image, where each pixel vector has been mapped to $\mathbb{D}^3$. However, the constraint is still given with respect to the original input image. To account for this, we can choose the step size $\alpha$ such that
\begin{equation}
    ||\tilde{\bfx} - \bfx||_2 = ||\log_{\mathbf{0}} (\exp_{\bfx} (\alpha \nabla_\bfx J(\theta, \bfx, y))) - \bfx||_2 = \epsilon,
\end{equation}
where $\tilde{\bfx}$ represents the generated adversarial sample in the image space. This value of $\alpha$ can be found using Newton's method.

The results of the comparison are shown in Figures \ref{fig:rfgsm_attacked_model_performance} and \ref{fig:rfgsm_vs_fgsm}. While there are some slight differences, the results are very similar to the results in Figures \ref{fig:fgsm_attacked_model_performance} and \ref{fig:misclass_mat_fgsm}. This is likely due to the fact that the perturbations are on such a small scale that the hyperbolic space is approximately Euclidean at this level. This shows that accounting for the geometry when performing the attack does not negate the discrepancy in behaviour of the two types of models. Somehow, the different geometries cause the models to learn different patterns, each with its own weaknesses and strengths. 

\vspace{-8pt}

\section{Conclusion}
\label{sec:conclusion}
\vspace{-5pt}
In this paper we proposed hyperbolic variants of the frequently used FGM and PGD adversarial attacks. The goal of these attacks is to improve adversarial performance by accounting for the geometry of the network. In the toy example of Section \ref{sec:toy_example} we have seen that these attacks help in certain cases, but not all. Furthermore, in Section \ref{sec:hyperbolic_models} we have seen that existing attacks perform differently versus Euclidean and hyperbolic networks, with hyperbolic networks being slightly more robust. Accounting for the geometry by applying the proposed hyperbolic attacks to the hyperbolic models did not remove any of the observed differences, demonstrating that the choice of geometry leads a model to learn different patterns with different strengths and weaknesses. As non-Euclidean geometry becomes increasingly pertinent to deep learning, future research into the resulting vulnerabilities and their origins will be of critical importance for ensuring adversarial robustness.

\vspace{-8pt}

\section*{Acknowledgments}
\label{sec:ack}
\vspace{-5pt}
Max van Spengler acknowledges the University of Amsterdam Data Science Centre for financial support. Jan Zah\'{a}lka's work has received funding from the European Union’s Horizon Europe research and
innovation programme under grant agreement No. 101070254 CORESENSE; and was co-funded by the European Union under the project ROBOPROX (reg. no. CZ.02.01.01/00/22\_008/0004590).

\bibliographystyle{splncs04}
\bibliography{main}
\end{document}